\documentclass{ieeeaccess}
\usepackage{cite}
\usepackage{amsmath,amssymb,amsfonts}
\usepackage{graphicx}
\usepackage{textcomp}

\usepackage{tabularx}
\usepackage{gensymb}

\usepackage{algorithm}
\usepackage{algpseudocode}

\def\BibTeX{{\rm B\kern-.05em{\sc i\kern-.025em b}\kern-.08em
    T\kern-.1667em\lower.7ex\hbox{E}\kern-.125emX}}
\begin{document}
\history{Date of publication 20.05.2024.}
\doi{10.1109/ACCESS.2017.DOI}

\title{Dynamic Line Rating using Hyper-local Weather Predictions: A Machine Learning Approach}
\author{\uppercase{Henri Manninen}\authorrefmark{1}, \IEEEmembership{Member, IEEE},
\uppercase{Markus Lippus\authorrefmark{2}, and Georg Rute}.\authorrefmark{2}}
\address[1]{Department of Electrical Power Engineering and Mechatronics, Tallinn University of Technology (e-mail: henri.manninen@taltech.ee)}
\address[2]{Grid Oracle OÜ, Tallinn, Estonia}

\markboth
{Henri Manninen \headeretal: Preprint}
{Henri Manninen \headeretal: Preprint}

\corresp{Corresponding author: Henri Manninen (e-mail: henri.manninen@taltech.ee)}

\begin{abstract}
Dynamic Line Rating (DLR) systems are crucial for renewable energy integration in transmission networks. However, traditional methods relying on sensor data face challenges due to the impracticality of installing sensors on every pole or span. Additionally, sensor-based approaches may struggle predicting DLR in rapidly changing weather conditions. This paper proposes a novel approach, leveraging machine learning (ML) techniques alongside hyper-local weather forecast data. Unlike conventional methods, which solely rely on sensor data, this approach utilizes ML models trained to predict hyper-local weather parameters on a full network scale. Integrating topographical data enhances prediction accuracy by accounting for landscape features and obstacles around overhead lines. The paper introduces confidence intervals for DLR assessments to mitigate risks associated with uncertainties. A case study from Estonia demonstrates the practical implementation of the proposed methodology, highlighting its effectiveness in real-world scenarios. By addressing limitations of sensor-based approaches, this research contributes to the discourse of renewable energy integration in transmission systems, advancing efficiency and reliability in the power grid.
\end{abstract}


\titlepgskip=-15pt

\maketitle

\section{Introduction} 

\PARstart{E}{lectrical} power systems are undergoing rapid changes due to ambitious climate goals that aim to integrate Renewable Energy Sources (RES) into the grid. For instance, the European Union targets generating 40\% of its electricity from RES by 2030 \cite{EU_Commission_RE_Directive_2021}. To achieve this energy transition cost-effectively, additional power transmission capacity is essential \cite{Child2019,Trondle2020, Joskow2020, Glaum2023}, but building new overhead lines (OHL) is expensive and time-consuming. Although expanding or upgrading power lines can increase transmission capacity, there is significant untapped potential in optimizing the capacity of existing high-voltage OHLs\cite{IRENA2020} by allowing them to operate at higher current densities and temperatures to increase the line’s maximum power flow. 

In thermally-limited OHLs, the maximum power that can be transmitted is limited by the maximum allowable temperature of the conductor, which should
not be exceeded to avoid overheating and to ensure safety distances due to extensive sagging. The flow of electricity heats up the conductors, while the rate of cooling depends on weather conditions. This has been observed and studied since the 1960's by different research groups in the USA such as House\&Tuttle \cite{Houseandtuttle} and used mainly for monitoring purposes. In the last decade Dynamic Line Ratings (DLRs) have been widely used by utilities and reported implementations of DLR have demonstrated an increase of transmission capacity of between 10 to 30 percent on average an annual basis \cite{Lai2022} and over 70\% increase for 95\% of time \cite{8269366}.

Dynamic Line Rating technologies usually fall into two broad categories, indirect methods and direct methods. Direct DLR approaches such as explained in \cite{KANG2020102086} depend on installing physical sensors to measure the properties of the actual conductor to determine line temperature. In the case of direct DLR technologies, ideally, each span should be monitored to be sure that there will be no clearance violation, but it is not economically viable. Since national power grids typically cover thousands of kilometers, it is not economically reasonable to install sensors on all spans. Instead, sensors are typically placed on the so-called critical spans which limit the capacity of the entire line. To identify critical spans, extensive power line modeling is undertaken based on historic or modelled weather data to identify the locations that are most likely to overheat or breach ground clearance limits as reviewed in \cite{Lai2022}. It has also been noted that when monitoring the whole OHLs, hotspots are not constant and they change based on wind speed and directions \cite{elevation_dlrZ2022107338, Karimi2018,Oncor2013}.

At high current levels, the temperature of the OHL conductor varies primarily through the effect of wind cooling and ambient temperature along the line route enabling to use indirect methods for conductor temperature calculation. Maximum allowed current or conductor temperature calculations are usually based on IEEE \cite{IEEE738} or CIGRE \cite{Cigre207, Cigre601} standards, where steady-state and dynamic-state methodologies are well explained. The advantage of using indirect approach is the independence from physical equipment, but it also relies mainly on meteorological observations \cite{DLr_forecasting}. In \cite{Molinar2019} the weather data was of granularity of 1km x 1km and in \cite{Matus2012} actual weather measurements from up to 20km away were used. As a way to overcome the lack of detail in wind data, in \cite{osti_1895360} computational fluid dynamics (CDF) was used to model the wind conditions over the length of a power line to assess the optimal spacing of sensors. The wind park production was modelled using CFD in \cite{CDF_NAKAO2019103970,CDF_wind_turbine}. While CDF is providing accurate results, it comes with a cost of significant computational resources. Existing numerical weather predictions can be accurate enough in flat and open areas, but may fail in complex terrain such as mountains or inside forests, where conductor temperatures can vary by 10 to 20 degrees due to variations in weather \cite{Seppa1993}. The effect of sheltering by trees, buildings or terrain may reduce the wind speeds by at least half of that recorded at open sites typical of meteorological sites \cite{KANG2020102086} leading to inaccuracies. Since power lines do go through such regions it is necessary to improve weather, and especially wind, forecasting in complex terrain to maximize the benefits DLR and safety of OHLs, especially in places where conductors are below canopy heights \cite{Phillips2017}. 

Numerical Weather Prediction (NWP) models are fundamental in weather forecasting, amalgamating data from various sources such as satellites, weather stations, and balloons. These models, when coupled with Global Climate Models (GCM), provide forecasts for atmospheric parameters. However, the spatial resolution of GCMs is constrained by computational demands, exemplified by ECMWF's \cite{ECMWF-IFS} medium-range model operating on a 9x9km grid, often inadequate for many applications. To achieve finer resolutions, both dynamic \cite{Dickinson_Errico_Giorgi_Bates_1989} and statistical \cite{stat-downscaling} downscaling methods are employed. Dynamic downscaling, exemplified by regional weather models (RWM), focuses on specific areas, enabling resolutions up to 2.5km. Conversely, statistical downscaling establishes relationships between model outputs and observed values, relying on historical data. However, both approaches face limitations. For tasks such as determining the maximum ampacity for OHLs, resolutions around 10m are imperative, posing challenges for existing downscaling methods. Dynamic downscaling becomes economically unviable due to escalating resource demands with increased resolution, while statistical methods are constrained to specific locations, overlooking terrain complexities crucial for precise evaluations.

This paper introduces a novel methodology that leverages hyper-local weather predictions to obtain high-resolution weather prediction for DLR using machine learning to combine the best knowledge from NWPs with location specific parameters such as terrain, elevation and vegetation. The proposed methodology also focuses on providing confidence intervals (CI) for wind predictions allowing grid operators to improve risk management by using CIs, quantile probabilities and other statistical quantities for DLR predictions. The proposed method is capable of covering large areas with the granularity of a single span, delivering reliable DLR estimations for up to 48 hours with corresponding confidence intervals. By focusing on span-level detail, this method helps to determine the actual limitations of each span and to prevent overheating of OHLs.

This paper is structured into four main sections. The first section (\ref{Methodology}) provides background information on the developed DLR assessment. The second section (\ref{Sec:Wind}), the hyper-local weather prediction model and its results are described. In the third section (\ref{Sec_Casestudy}) the implementation of using hyper-local weather prediction on DLR is explained and section \ref{Sec_Casestudy} presents a case study illustrating the methodology's effectiveness, demonstrated through its application to a single OHL in Estonia.

\section{Dynamic line rating at a granular span level} \label{Methodology}  

The proposed methodology aims to address the challenge of longitudinal temperature variation along overhead lines (OHLs) due to varying local weather conditions, which can lead to substantial temperature discrepancies along the conductor. This variation is poorly equalized axially, resulting in significant temperature differences, especially at high current densities. Based on  [54] differences of 50ºC in conductor temperature have been
reported within a single partially sheltered span at very high current densities. The proposed methodology is designed to calculate Dynamic Line Rating (DLR) at a granular span level, maximizing efficiency. In addition to reliable weather parameters, accurate data about the OHL conductor, including maximum allowed temperature for each span, is essential for trustful results.

The methodology involves several steps outlined in the Algorithm \ref{alg:DLR_Calculation}. The process begins by iterating through time periods, followed by iterating through all spans along the OHL. At each span iteration, essential data inputs are gathered, including conductor parameters and allowed span temperatures. Concurrently, hyper-local weather predictions are obtained, leveraging Numerical Weather Predictions (NWP) and Digital Terrain Models (DTM) and Digital Surface Models (DSM) and are elaborated in Section \ref{Sec:Wind}. The hyper-local weather predictions and terrain data are then utilized to calculate confidence intervals (CIs) for DLR assessments. These CIs provide insights into the reliability of the DLR predictions, enabling grid operators to make informed decisions. 

Next, DLR calculations are performed based on the gathered data, considering the temperature variations along the OHL caused by changing local weather conditions. The calculated DLR values are stored for each span, facilitating further analysis and comparison. After completing the DLR calculations for each span within the time period, the process iterates to the next span and time period, ensuring a comprehensive assessment of DLR across the entire OHL network. Finally, data aggregation occurs to produce a single DLR value for each OHL, taking into account all individual span calculations. This aggregated DLR data provides valuable insights for optimizing transmission system operations and improving grid reliability. The DLR for the OHL is calculated as the minimum of the DLRs for each span for each time period.

\begin{algorithm}
\caption{DLR Calculation Process}
\label{alg:DLR_Calculation}

\begin{algorithmic}[1]
\State \textbf{Start}
\State Iterate over time periods
    \For{each time period}
        \State Iterate over spans
            \For{each span}
                \State Gather conductor parameters 
                \State Gather allowed span temperatures
                \State Gather NWP and terrain data     
                \State Perform hyper-local weather prediction 
                \State Calculate CI for weather predictions
                \State Calculate DLR 
                \If{more spans}
                    \State Move to the next span
                \Else
                    \If{more time periods}
                        \State Move to the next time period
                    \EndIf
                \EndIf
                \State Store DLR
            \EndFor
    \EndFor
\State Calculate $DLR=\min\left( \text{DLR}_{\text{span}_1}, \text{DLR}_{\text{span}_2}, \ldots, \text{DLR}_{\text{span}_n} \right)$
\State \textbf{End}
\end{algorithmic}
\end{algorithm}

\section{Hyper-local Weather Prediction} \label{Sec:Wind}

The hypothesis behind this approach is that detailed information about the geography of the immediate environment allows models to produce more precise forecasts in these settings. This hypothesis relies on the model's ability to learn relevant features from the data and to correlate different sources of information effectively.

\subsection{Weather Forecasting}
Weather forecasting relies heavily on numerical weather prediction (NWP) models, which integrate data from various sources such as satellites, weather stations, and weather balloons. These data inputs are processed using global climate models (GCM) to generate predictions for atmospheric weather parameters at both ground level and higher altitudes. However, one significant challenge with GCMs is their computational cost, which makes high spatial resolution impractical. For instance, the European Centre for Medium-Range Weather Forecasts (ECMWF) utilizes a high-resolution forecasting model with a 9x9 km grid, which, while useful for many applications, falls short in scenarios requiring finer resolution, such as calculating the maximum ampacity of OHLs.
To achieve higher resolution forecasts, two main methods are employed: dynamic and statistical downscaling.

\subsubsection{Dynamic Downscaling}
Dynamic downscaling involves using regional weather models (RGM) such as HARMONIE-AROME\cite{arome}, which cover smaller areas and use GCM outputs for initial and boundary conditions. This approach allows RGMs to operate at much higher resolutions, commonly around 2.5 km, but potentially higher, in the case of HARMONIE-AROME, thereby reducing the computational load while maintaining accuracy over a limited area \cite{Dickinson_Errico_Giorgi_Bates_1989}.
In Northern Europe, the Norwegian Meteorological Institute is the key player in running and disseminating results from the MetCoOp Ensemble Prediction System \cite{metno}, a local, high resolution NWP system based on HARMONIE-AROME. 

\subsubsection{Statistical Downscaling}
Statistical downscaling, on the other hand, relies on statistical relationships between model outputs and observed values. This method’s accuracy depends on extensive historical data and can be applied directly to GCM outputs or preferably to RGM outputs when available for the region. However, statistical methods have limitations, especially when extremely high resolutions, e.g. 10 meters, are required and when specific terrain and vegetation characteristics significantly impact weather parameters \cite{stat-downscaling}.

\subsubsection{Machine Learning Approach for Hyper-local Forecasting}
To address these limitations, especially for applications like determining the maximum ampacity of OHLs, a machine learning approach is approached. This method assimilates relevant data and learns to downscale weather predictions by considering location-specific characteristics without being tied to particular locations. This innovative approach enables the generation of hyper-local weather predictions, providing fine-grained forecasts essential for precise applications.

\subsubsection{Wind Velocity Components in Weather Forecasting}
The $u$ and $v$ components of wind velocity are fundamental in describing wind direction and speed in meteorological models. In this system the $u$ component represents the east-west wind component and the $v$ component represents the north-south wind component. Using these components, wind speed ($w_i$) and wind direction ($\theta_i$) are calculated according to \ref{wind_speed} and \ref{wind_direction} respectively.

\begin{equation}
\label{wind_speed}
w_i = \sqrt{u_i^2 + v_i^2}
\end{equation}

\begin{equation}
\label{wind_direction}
\theta_i = \arctan{\left(\frac{v_i}{u_i}\right)} + \pi
\end{equation}

Representing wind as components is crucial for NWP models and meteorological analyses as they simplify the representation of wind patterns. By breaking down wind velocity into $u$ and $v$ components, meteorologists can more effectively understand and predict wind behavior, facilitating better insights into the movement and evolution of weather systems over time.

\subsection{The Hyper-local Weather Prediction Model}
The inputs to the model comprise the outputs of the NWP models and a number of features describing the terrain collected by measurements from satellites and airplanes. As the focus is to predict ambient temperature, wind speed and direction, a set of 5 surface boundary layer weather parameters is used - air temperature at 2m, wind u component, wind v component, pressure at sea level and total precipitation. Also four data sets related to local topography - 1 arcsecond DEM (digital elevation model), land use and land cover, 1m DTM (digital terrain model), 1m DSM (digital surface model) and their derivatives are used. The data are processed into squares around the point of interest resulting in \(x * y * c\) matrices where \(x\) and \(y\) are the side lengths of the area in pixels and \(c\) is the number of input features from the area. As the LiDAR, satellite and weather data have different resolutions, the features from these separate sources are handled separately by the model downstream.

\Figure[t!](topskip=0pt, botskip=0pt, midskip=0pt)[width=0.99\linewidth]{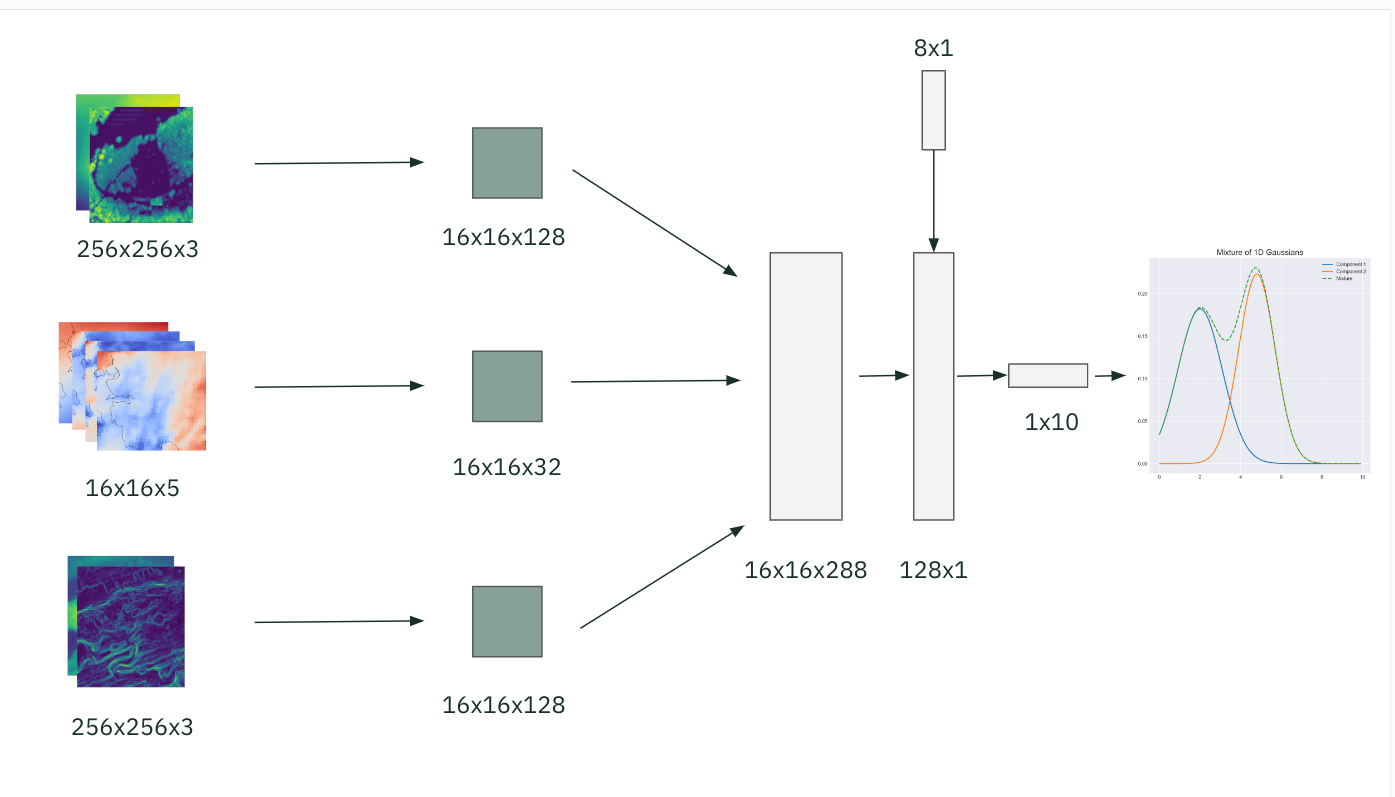}
{The basic structure of the model. Input features are processed by separate input branches and the extracted features processed together to learn interactions between the features. The estimates for the distribution parameters of the wind components are based on the aggregate of all inputs.\label{fig:model}}

The goal of the model is to learn a function that maps the low resolution weather parameters onto specific locations, taking into account the local topography of the point of interest. To do that the model must learn how the topography influences these parameters and the various interactions between the inputs. The task is complicated by the input weather parameters not being actual measurements but model outputs, containing noise.
As target variables, the actual measurements of wind speed and direction from weather stations are used. Measurements are aggregated to hourly frequency, taking the average of measurements made during the last 10 minutes of every hour. These measurements are transformed into \(u\) and \(v\) components of the wind, which lend themselves well for machine learning purposes.

To model and quantify uncertainty in the predictions, the model is not built to provide specific values for the wind components, but parameters for a distribution of possible values. As the wind components are Gaussian they can be modelled using a normal distributions, but being correlated, using a joint distribution makes more sense. The initial experiments showed that a Gaussian Mixture Model (GMM) works well in terms of modelling capability and adjusting to different conditions. Considering wind components as being sampled from distributions with different parameters depending on the conditions enables the model to learn how the conditions influence the distribution parameters and the mixing distribution. It also allows the model to make better use of varied measurement data, generated by instruments of differing reliability and accuracy, enabling the model to take these details into account.

Deep learning has been extensively utilized for working with remote sensing data. However, due to the varied nature of the data and objectives within this field, which is less popular than fields dealing with common RGB image processing, there are no well-established methods for learning from these data or widely adopted task-specific architectures outside of segmentation.

Although most inputs in remote sensing can be treated as 2D or 3D matrices and many computer vision approaches can be applied, experiments indicate that specific architectures and pretrained models are often not very useful. This is primarily due to the following reasons:
1. The source data is very different from common images both in meaning and in form. Unlike RGB images elevation ranges from negative values to thousands of meters, represented in floating point numbers, with no defined number of channels.
2. The input data varies in scale and nature. Treating it as a single input does not work very well.

\subsubsection{Data Preparation}

This means that the inputs are different in shape and range, rendering most pretrained models useless unless very specific data preparation steps are used, which could limit the capability of the model and risk information loss.
To handle these issues, a multi-input model is used to processes data at different scales separately, extracts features and combines the extracted features at similar scales. At the same time as geographically distinct samples are scarce - limited by the number of weather stations - the model complexity must be kept low to learn relevant features and not to overfit.

In total the model has four separate inputs:
\begin{enumerate}
    \item Low-resolution remote sensing data - satellite measurements and their derivatives.
    \item High-resolution remote sensing data - LiDAR measurements and their derivatives.
    \item NWP data - forecasts from either a local or a global NWP model, depending on the region and availability.
    \item Metadata - month and hour, forecast lead time, general geographical region and specifics of the NWP data used.
\end{enumerate}

Even in a split like this, the source data often varies in resolution and projections, especially in the case of forecast data, where resolutions can differ by an order of magnitude. In this case, data is down- or upsampled to a shared resolution and geographically aligned. For example, if the satellite sourced DEM uses WGS84 projection at 1 arcsecond resolution and Land Use map uses EPSG:3857 at a resolution of 10m, we reproject both into a local equal area projection and downsample Land Use data to match the DEM.
General location information is provided at a precision of approximately 1 degree in EPSG:4326 to avoid the model overfitting to a location, but still provide a general area. This is important due to the variability in data sources - LiDAR data is collected and processed differently, resulting in variability in measurement errors and processing artifacts, while weather models differ in methodology, output resolution and geographical biases.

Inputs 1 and 2 are downsampled using modified ResNet blocks \cite{2015resnet} to learn local features from the data and reduce the size of the larger inputs. CBAM \cite{cbam} a specific sample. When all inputs are sampled to a similar size, the intermediary feature maps are concatenated channel-wise and fed through MaxViT blocks \cite{tu2022maxvit} to learn interactions between the inputs. Finally the feature maps are pooled, flattened and concatenated with input 4. This is processed by an MLP, that produces the parameters for the GMM. During training the model is evaluated by the goodness of fit of the distribution and the observed variables using negative log-likelihood. 

When trained, the model provides probability distributions for wind components. These are further processed to provide confidence intervals for wind speed and direction.

\Figure[t!](topskip=0pt, botskip=0pt, midskip=0pt)[width=0.99\linewidth]{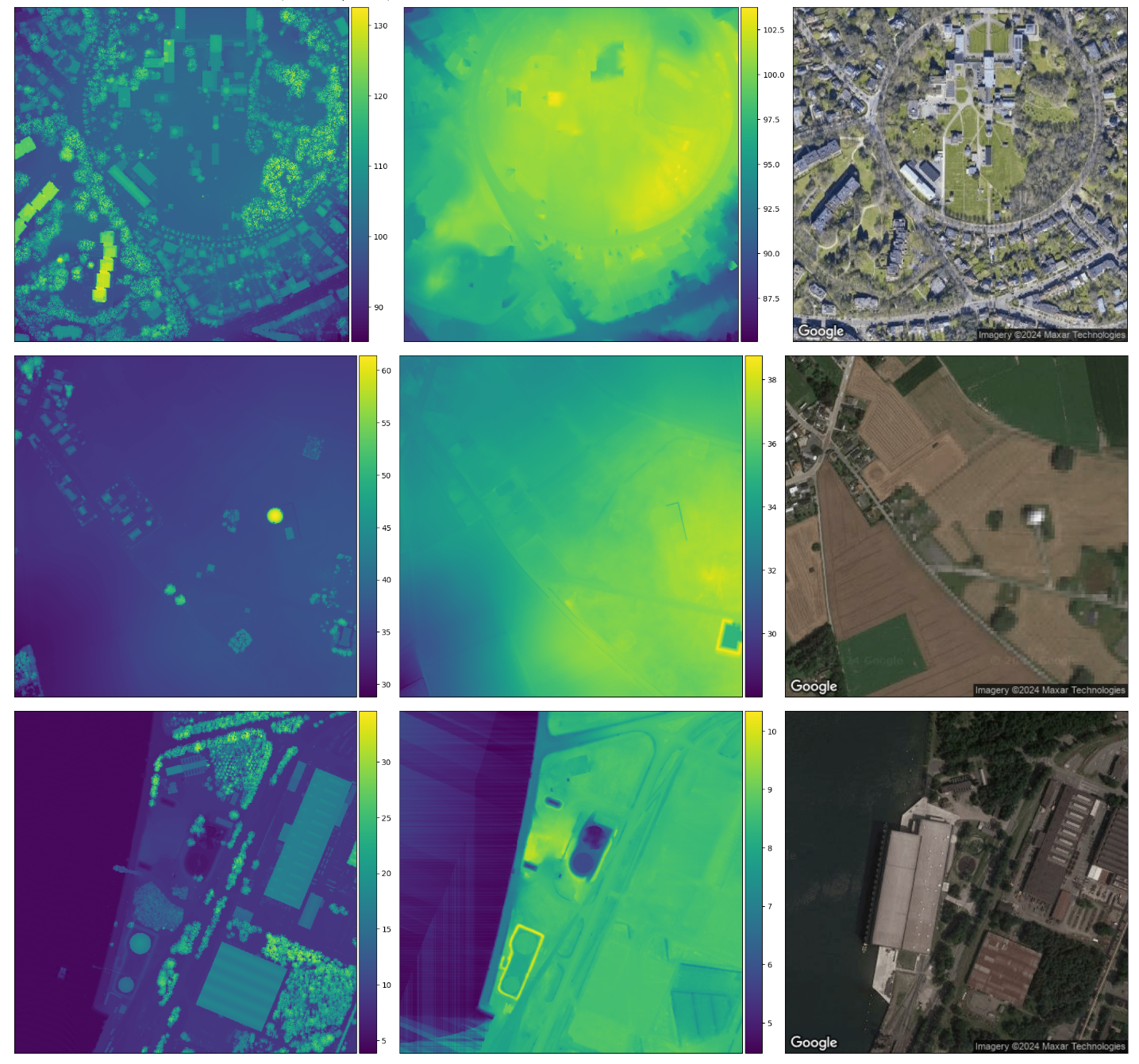}
{Inputs to the model, where left is DTM, middle DSM and right the actual location. \label{fig:dtm_dsm_ortho}}

\subsection{Data}

The data set used to train this model comprises measurements from 711 weather stations around Europe with a maximum range in time from 2018-2024 FIGURE. This comes to about 15 million samples, where a minimal sample consists of:
1. Wind components of the actual measured values - the target variables.
2. NWP output - forecast for a point in time, not reanalysis.
3. LiDAR data - high resolution terrain and surface models in the area.
4. Satellite data - low resolution surface models and masks.

Geographic data varies in projections, resolutions and minute transformations. For every specific region the projection is selected to have minimal distortions to the data, while working with cartesian coordinates and an equal-area projection to be suitable for the model. Input data is also resampled to share the resolution, opting for minimal information loss. For example, all NWP inputs are upsampled to 1000m resolution.
For every set of measured data, the relevant region is extracted from the source data, centering on the location where the measurement was made. As the sources are processed separately by the model, different geographical ranges are used for the inputs by taking the full advantage of high resolution LiDAR data while having a wider view of the terrain of the area from satellite data.
While NWP data is minimally processed, additional features are generated from elevation maps for the model to learn from. As the amount of distinct topography samples is relatively small to the input size, it generates relevant features for the machine to learn from, leading to increase in performance. Slope and histogram equalization matrices  are produced from surface elevation and added to the elevation as channels.

The key features included in the dataset are summarized in Table. \ref{tab:data_description} and Table. \ref{tab:data_summary} presents a summary of the dataset distribution across the training, testing, and validation sets, including the actual number of observations in each subset. 

\begin{table}
\centering
\caption{Summary of Key Features in the Input Data}
\label{tab:data_description}
\begin{tabularx}{\columnwidth}{|X|X|X|}
\hline
\textbf{Feature} & \textbf{Description}                                               \\ \hline
Temperature      & Air temperature (in Kelvin)                                       \\ \hline
Pressure         & Atmospheric pressure (in hPa)                                      \\ \hline
Humidity         & Relative humidity (in percentage)                                  \\ \hline
U                & U component of the wind                                  \\ \hline
V                & V component of the wind   \\ \hline
DTM@1m        & Ground elevation at 1m resolution  - 256x256m                 \\ \hline
DSM@1m        & Surface cover elevation at 1m resolution - 256x256m                  \\ \hline
DEM@1arcsecond    & Surface elevation at 1 arcsecond resolution ~ 5632x5632m                  \\ \hline
Time                & Month and hour of the measurement   \\ \hline
Location                & Coordinates at 1 degree accuracy   \\ \hline
Forecast lead time                & Nr. of hours the forecast was made in advance   \\ \hline
NWP identifier                & ID for the forecasting model the forecast inputs are from   \\ \hline
\end{tabularx}
\end{table}

\begin{table}
\centering
\caption{Dataset Summary }
\label{tab:data_summary}
\begin{tabularx}{\columnwidth}{|l|X|}
\hline
\textbf{Dataset}   & \textbf{Number of Observations} \\ \hline
Training           & 10,031,766                      \\ \hline
Testing            & 1,407,202                       \\ \hline
Validation         & 3,874,804                       \\ \hline
Total              & 15,313,772                       \\ \hline
\end{tabularx}
\end{table}

\subsection{Location independence}
Neural networks are good at approximating complex functions, there is a real danger of the model overfitting to specific locations and working very well there, while being bad at generalization. The goal for this work is to be able to generate more accurate forecasts independent of the location to be able to cover all the possible locations for power lines. To do this, the inputs are selected to provide data to learn general patterns from and not give the model data to overfit to.
It's not possible to provide no information about the location itself, but as the topographical data is fairly large - 65536 input parameters per layer - it should be enough to avoid learning location specific patterns. To evaluate that the model is indeed learning to generalize, the data is split three:
\begin{enumerate}
    \item Training data - all samples from a stations up to the last year we have data for
    \item Validation data - the last year of observations for each station.
    \item Test data - stations not included in training and test data with all measurements from the locations. 
\end{enumerate}

The measured wind training, validation and testing data distribution is illustrated more precisely in Fig.\ref{fig:wind_direction_data} and Fig.\ref{fig:wind_speed_data}. As seen all three data sets are similar in terms of distributions.

\Figure[t!](topskip=0pt, botskip=0pt, midskip=0pt)[width=0.99\linewidth]{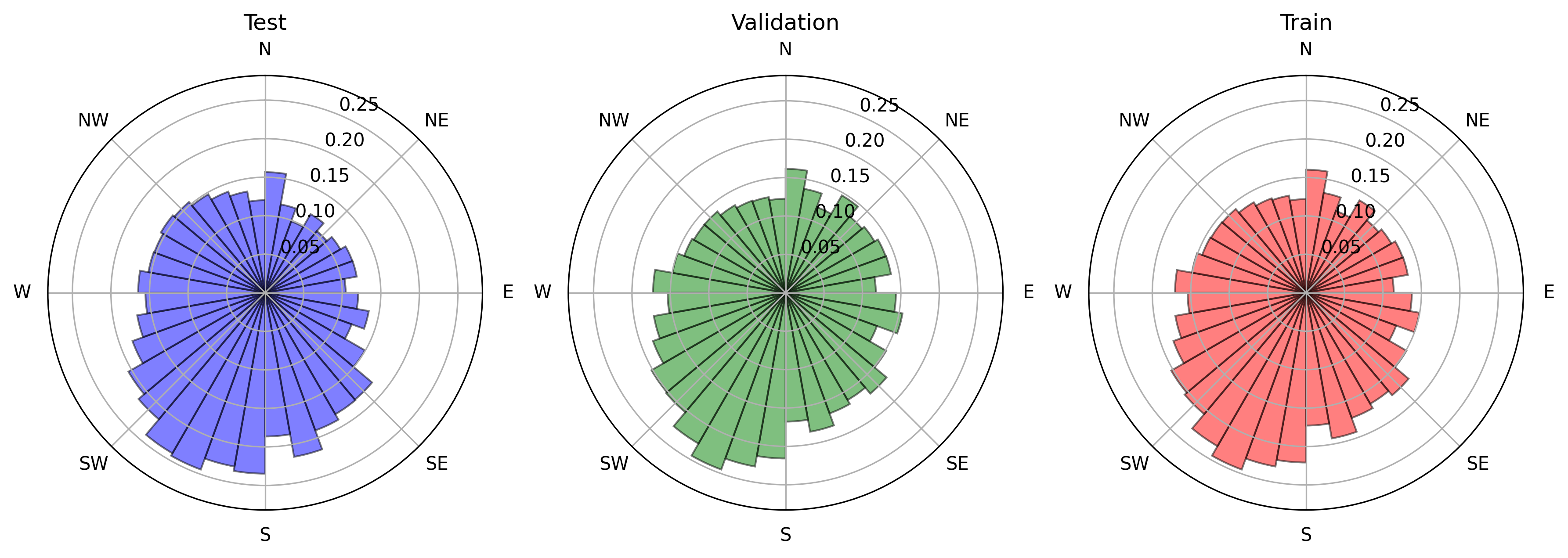}
{Measured wind directions for test, validation and train data.\label{fig:wind_direction_data}}

\Figure[t!](topskip=0pt, botskip=0pt, midskip=0pt)[width=0.99\linewidth]{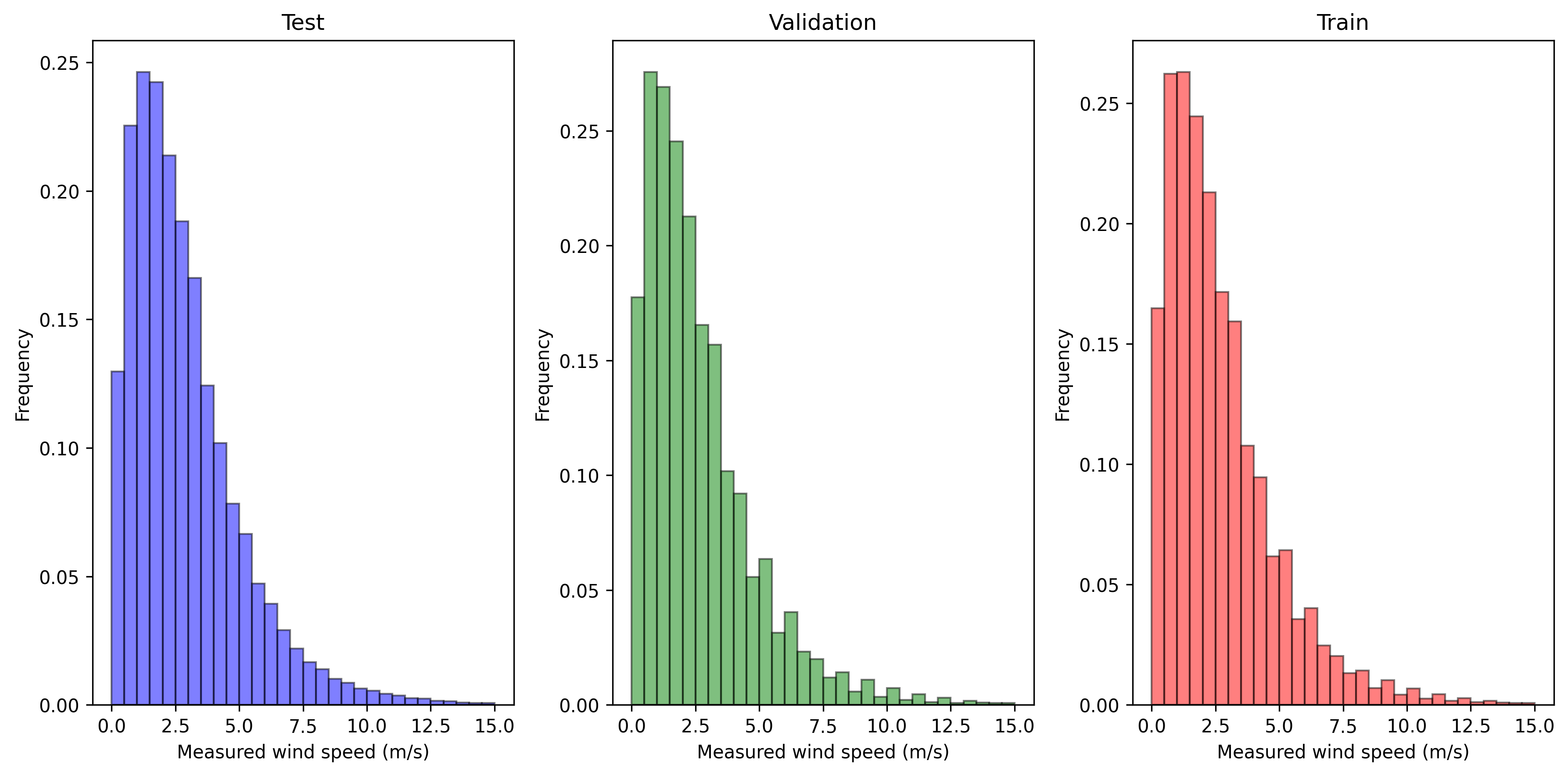}
{Measured wind speed for test, validation and train data.\label{fig:wind_speed_data}}

While it is expected that the model produces better results on the validation data, comprising stations it has seen earlier, if the difference in performance between validation and test become too large, then it can be assumed the model is overfitting to specific locations and losing its ability to generalize.

\Figure[t!](topskip=0pt, botskip=0pt, midskip=0pt)[width=0.99\linewidth]{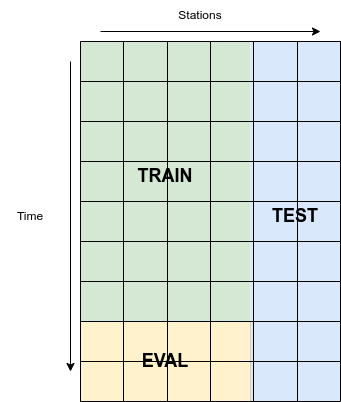}
{Data split}\label{fig:train_test}

\section{Weather Prediction Results and Confidence Intervals}
Wind speed, direction and ambient temperature  are pivotal parameters in DLR, yet they are often treated as constant values without consideration of their distribution. However, these measurements are prone to uncertainty stemming from various factors, including prediction errors, measurement inaccuracies, spatial variability, and atmospheric turbulence. Estimating confidence intervals (CIs) for wind speed and direction is essential for comprehending and forecasting weather patterns and for mitigating risks associated with overheating OHLs. In this sections shows results from \ref{Sec:Wind} and a method for estimating CIs through Monte Carlo simulation and fitting statistical distributions is proposed.

\subsection{Prediction Results}
The weather prediction models are evaluated on a number of metrics to make sure that the model:
\begin{enumerate}
    \item Is accurate in comparison to baseline - the baseline being the input from the NWP models.
    \item Is not sacrificing either direction or speed accuracy in favor of the other.
    \item Is uniformly accurate in terms of locations ad wind speeds.
    \item Produces minimal amounts of uncertain predictions
\end{enumerate}

A comparison of the metrics can be seen on Fig. \ref{fig:windmodel_error_all} and
Fig. \ref{fig:windmodel_error_low}. Our model is significantly more accurate than the NWP models regarding wind speed both in the case of validation and test data sets and both in terms of all observations and only low wind speed cases. The model does not overfit to specific sites, but is at about the same level as NWP regarding wind direction.

\Figure[t!](topskip=0pt, botskip=0pt, midskip=0pt)[width=0.99\linewidth]{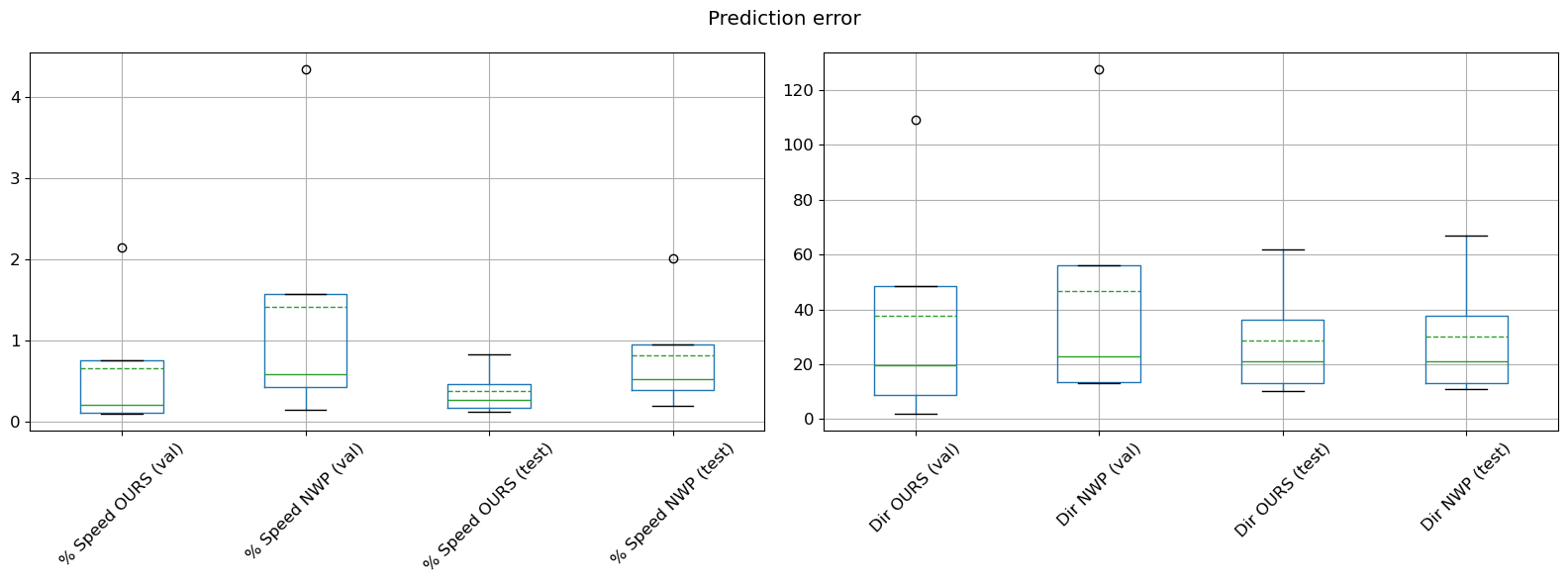}
{Errors for wind speed and direction on validation and test data for all samples. \label{fig:windmodel_error_all}}

\Figure[t!](topskip=0pt, botskip=0pt, midskip=0pt)[width=0.99\linewidth]{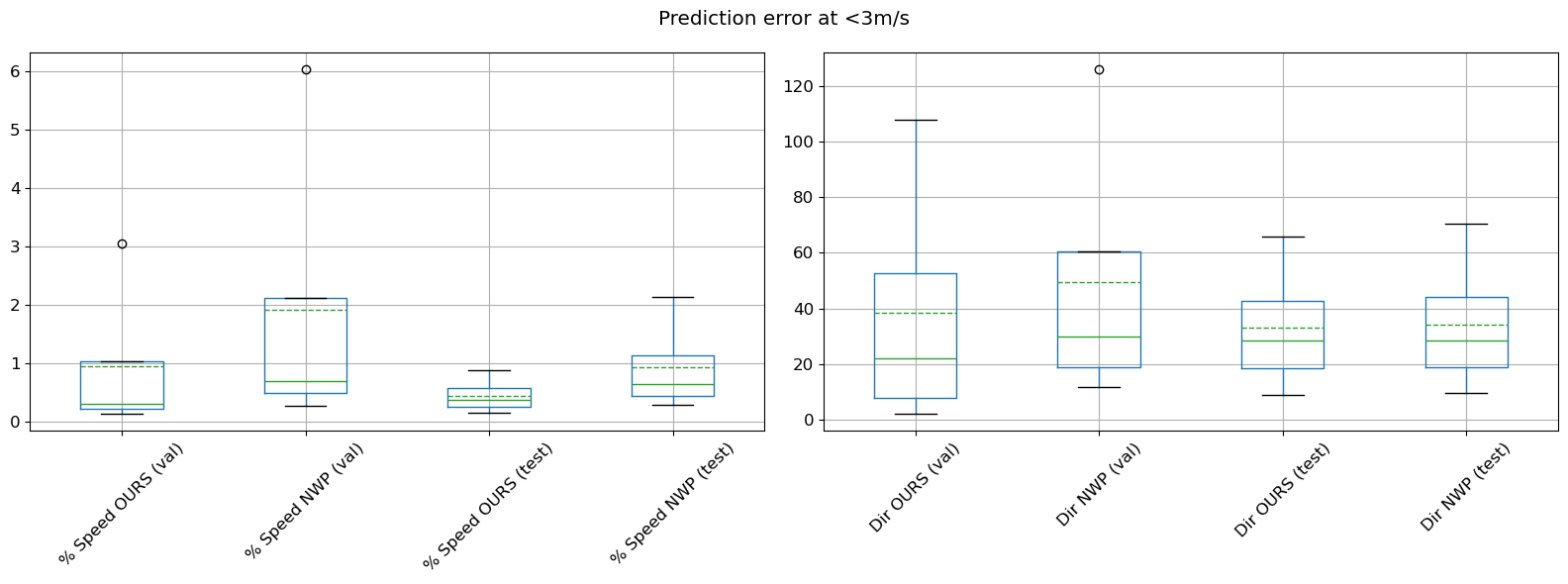}
{Errors for wind speed and direction on validation and test data for samples where wind speed is below 3 m/s.\label{fig:windmodel_error_low}}

Fig. \ref{fig:wind_direction_errors} shows the wind direction error using MAE and RMSE are similar for both, using the proposed methodology and using NWP.  Fig. \ref{fig:wind_speed_errors} presents that the proposed methodology decreases both MAE and RMSE by around 35\%.

\Figure[t!](topskip=0pt, botskip=0pt, midskip=0pt)[width=0.99\linewidth]{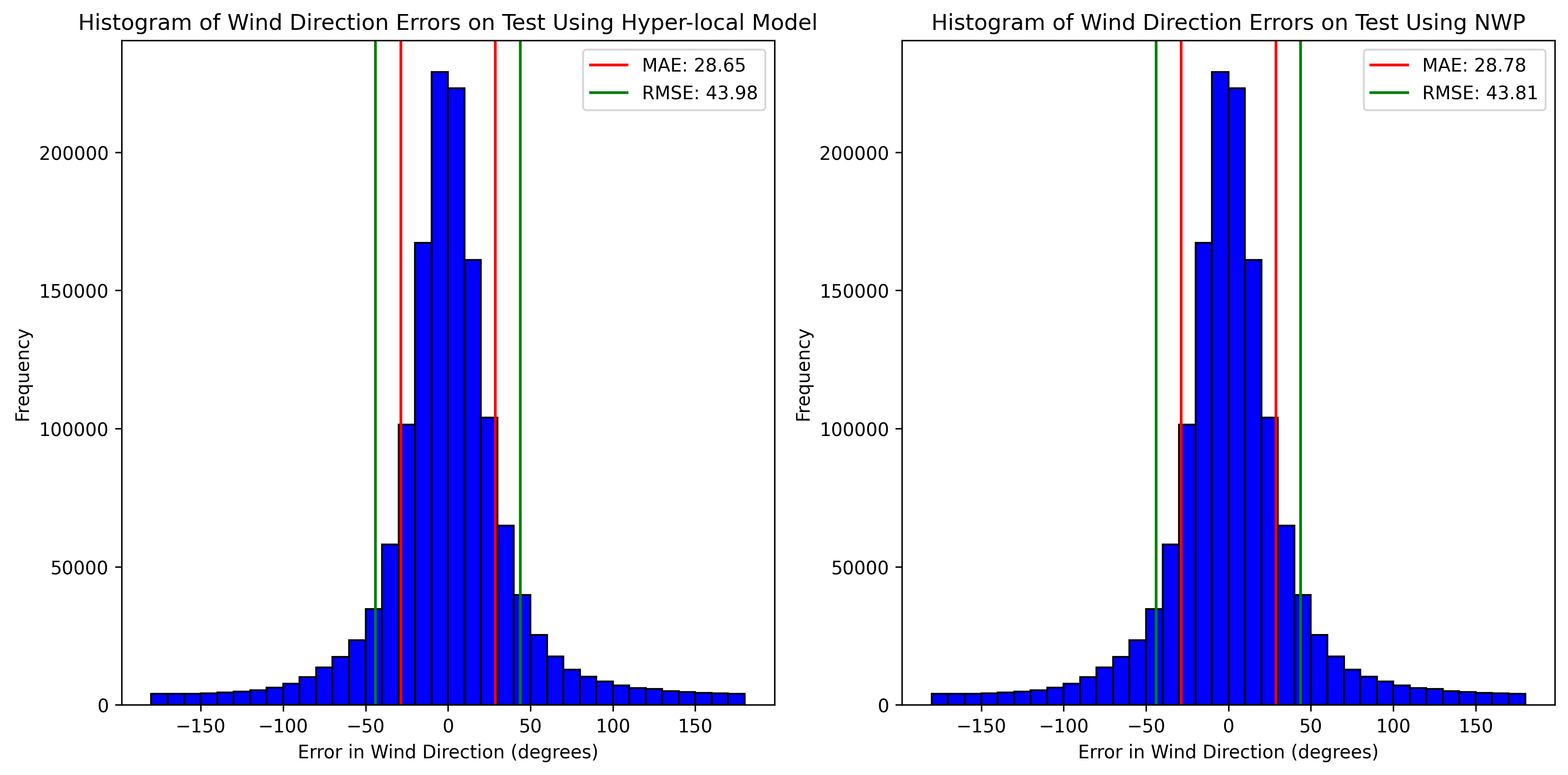}
{Wind direction errors with MAE and RMSE.\label{fig:wind_direction_errors}}

\Figure[t!](topskip=0pt, botskip=0pt, midskip=0pt)[width=0.99\linewidth]{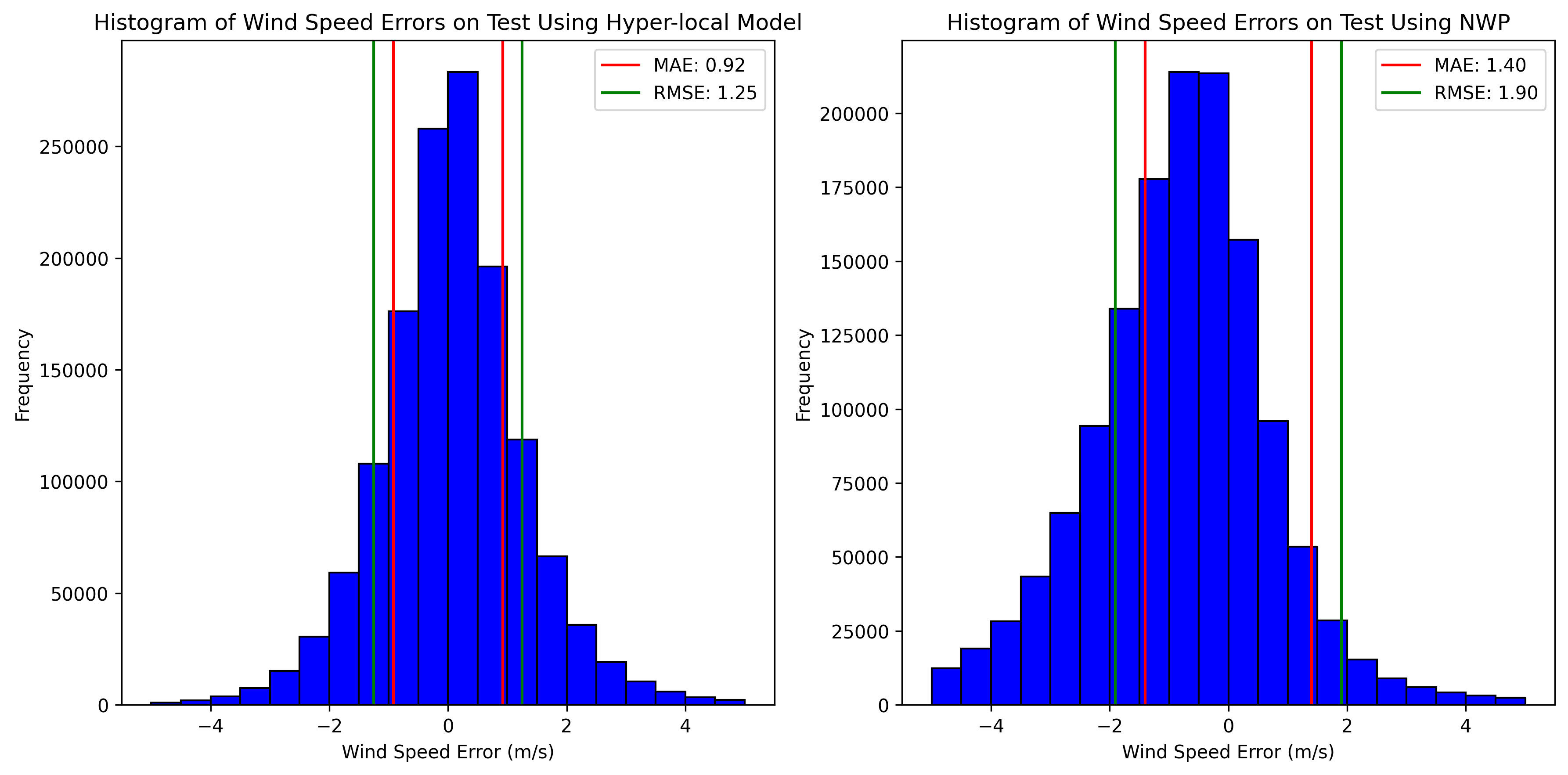}
{Wind speed prediction errors with MAE and RMSE.\label{fig:wind_speed_errors}}

\subsection{Confidence Intervals}

The outputs of the proposed hyper-local weather prediction model yield the $u$ and $v$ components of wind and temperature with mean and standard deviation values, following a normal distribution for each prediction. However, once wind speed and direction are calculated using these components, the normal distribution may not accurately capture their characteristics. Wind speed, for instance, is never negative and typically exhibits a skewed distribution with a long tail. Likewise, wind direction is a circular value ranging from 0 to 360 degrees. Consequently, more suitable distribution functions are warranted to account for these nuances. The ambient temperature and its errors are implemented with normal distribution.

\subsection{Distribution Fitting for Wind Speed and Direction}

The hyper-local weather prediction model outputs wind components with normal distribution. However, in wind modelling it is recommended to use log-normal, truncated normal or mixture of both for wind speed distributions \cite{lognormal_Baran_2015} and finite mixture of von Mises distributions \cite{wind_speed_and_preiciton_analysis} for wind direction. Two-Parameter probability distributions \cite{wind_speed_and_direction_pdf} are used in wind power modelling when the location of wind park is known. As the aim of the paper is to predict DLR for unseen locations then log-normal distribution is used for the wind speed and von Mises for wind direction to make the model more general. To increase the reliability of results, it is recommended to analyse more detailed distribution fitting  methodologies in further studies.

The proposed method utilizes Monte Carlo simulation and statistical distribution fitting to generate random samples of wind components, compute derived quantities (wind speed and direction), and estimate their CIs. To generate random samples of wind components, the Monte Carlo simulation technique is used. Given the mean ($\bar{u}$, $\bar{v}$) and standard deviation ($\sigma_u$, $\sigma_v$) of wind components (eastward and northward wind speeds), random samples ($u_i$, $v_i$) are generated from normal distributions.

\begin{algorithm}
\caption{Monte Carlo Simulation and Distribution Fitting for Wind Components}
\begin{algorithmic}[1]
\Require Mean and standard deviation of wind components: $\bar{u}$, $\bar{v}$, $\sigma_u$, $\sigma_v$; Confidence Interval (CI)
\Ensure Confidence intervals for wind speed and direction
\State Initialize number of samples $N=1000$
\For{$i = 1$ to $N$}
    \State Generate $u_i \sim \mathcal{N}(\bar{u}, \sigma_u)$
    \State Generate $v_i \sim \mathcal{N}(\bar{v}, \sigma_v)$
    \State Compute wind speed $w_i = \sqrt{u_i^2 + v_i^2}$
    \State Compute wind direction $\theta_i = \arctan\left(\frac{v_i}{u_i}\right) + \pi$
\EndFor
\State Fit log-normal distribution to $\{w_i\}$
\State Fit von Mises distribution to $\{\theta_i\}$
\State Calculate CI for wind speed using log-normal
\State Calculate CI for wind direction using von Mises
\State \Return CIs for wind speed and direction
\end{algorithmic}
\end{algorithm}

The proposed function generates random samples from normal distributions representing wind components and computes derived quantities using trigonometric transformations. Statistical distributions to the derived quantities of wind speed ($w_i$) and wind direction ($\theta_i$) are fitted. Specifically, a log-normal distribution is fitted to wind speed and a von Mises distribution to wind direction using maximum likelihood estimation. Using the fitted distributions, it is possible to calculate CIs for wind speed and direction.

\subsubsection{Confidence Intervals on Wind Attack Angle}

However, getting the minimal attack angle based on wind direction CI is more difficult as the bearing of each span can be random and using upper and lower bound of 360\degree circle requires a more detailed approach. The Algorithm \ref{alg:wind_dir} proposes a way to deal with wind direction confidence intervals on the minimal attack angle calculation. 

\begin{algorithm}
\caption{Calculate Minimal Attack Angle}
\label{alg:wind_dir}
\begin{algorithmic}[1]
\State Normalize wind directions to 0-360\degree.
\State Convert wind directions to radians.
\State Convert OHL directions to radians.
\State Compute the angular range of the wind direction CI.
\For{each span direction}
    \If{span direction within bounds}
        \State \textbf{return} the minimal angle = 0.
    \Else
        \State Calculate absolute circular differences.
    \EndIf
\EndFor
\State Determine the minimal angle by identifying the smallest absolute difference.
\State Adjust the minimal angle for circular space discontinuities.
\State \Return the minimal angle.
\end{algorithmic}
\end{algorithm}

Fig.\ref{fig:attack_angle_explanation} presents an example of wind direction mean value CIs and how the minimal attack angle is determined. As seen from the figure, the the minimal attack angle is affected by CI range, wind direction mean and OHL's span bearing. In this example, the minimal attack angle is determined by the upper bound of CI and is resulting an attack angle of 22\degree. 

\Figure[t!](topskip=0pt, botskip=0pt, midskip=0pt)[width=0.7\linewidth]{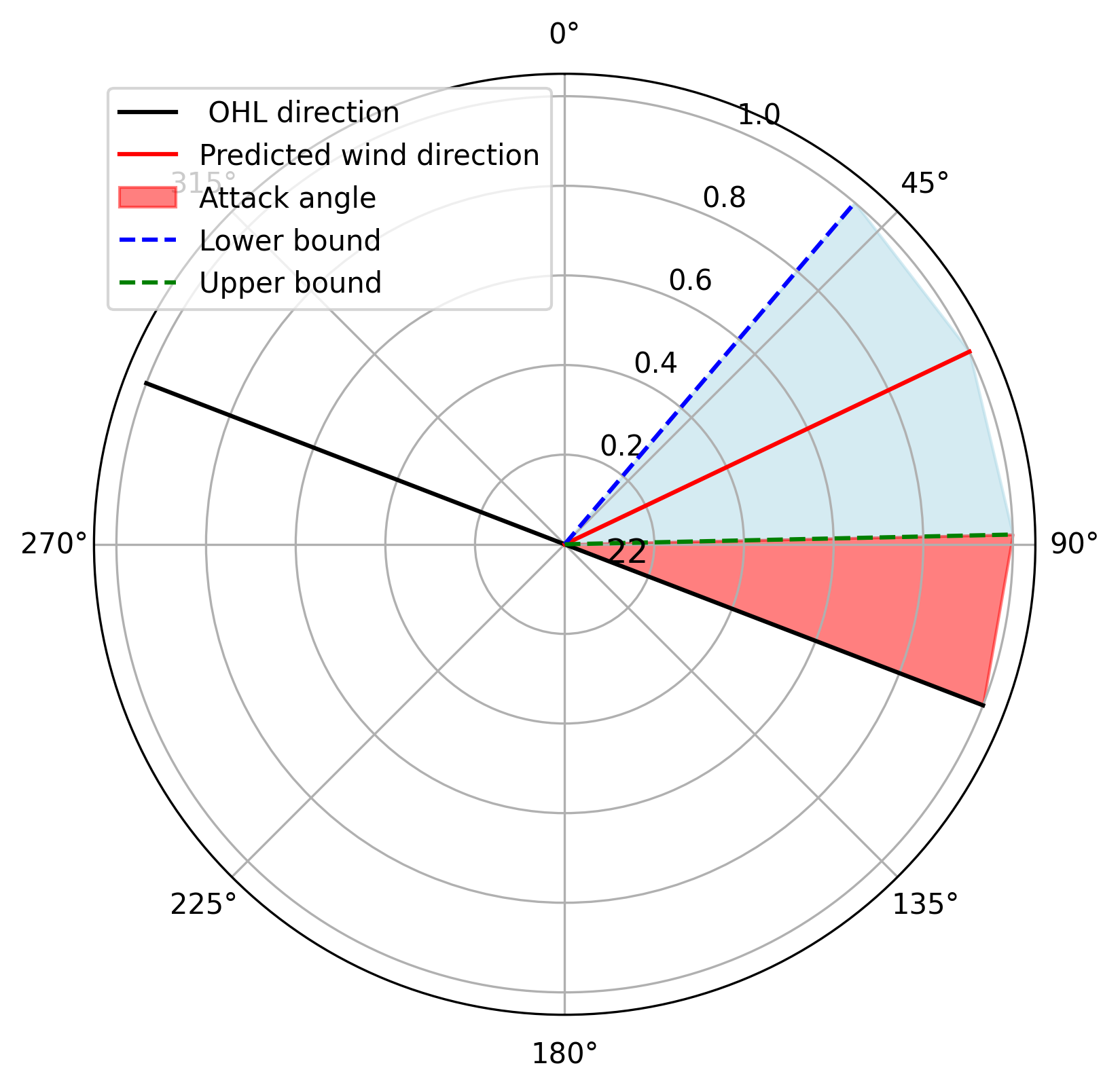}
{An explanation of the minimal wind attack angle and confidence intervals.\label{fig:attack_angle_explanation}}

\section{Dynamic line rating at a granular span level} \label{DLR}
In the paper, it is assumed that the conductor is in a steady-state, where the maximum current and temperature are both considered to be constant. Input parameters are also considered to be constant over the time period.

\subsection{Thermal Ratings of Overhead Lines} 
The calculation of thermal ratings of OHLs have been extensively discussed and improved over the past decades by CIGRE and IEEE. Two of the most widespread methodologies for the assessment of thermal ratings are CIGRE Technical Brochure 207 \cite{Cigre207} and IEEE standard 738 \cite{IEEE738}. They both have been widely used for almost 20 years to on determine the steady-state heat balance of conductors as described with \ref{steady_state}. CIGRE TB 207 only covers the thermal behaviour of overhead conductors at low current densities ($<1.5A/mm2$) and low temperatures ($<100\degree C$) for ACSR conductors. In CIGRE Technical Brochure 601 \cite{Cigre601} numerical and analytical models for both steady-state are improved and transient-state with examples on how to perform temperature tracking are added. 

\begin{equation}
\label{steady_state}
P_j + P_s + P_m + P_i = P_c - P_r - P_w
\end{equation}
where
\begin{align*}
&P_j \text{ is the Joule heating,} \\
&P_s \text{ is the solar heating,} \\
&P_m \text{ is the magnetic heating,} \\
&P_i \text{ is the corona heating,} \\
&P_c \text{ is the convective cooling,} \\
&P_r \text{ is the radiative cooling, and} \\
&P_w \text{ is the evaporative cooling.}
\end{align*}

In practice, the terms of the heat balance equation above the thermal state of OHLs depends mainly on ambient weather parameters such as wind speed and direction, ambient temperature, solar radiation, and on the electrical current flowing through it. Corona heating and evaporative cooling have a minimal impact are not used in the scope of this paper. The methodology used in this paper for the calculation of thermal ratings of OHLs is based on the CIGRE TB 601 \cite{Cigre601} with modifications in solar irradiance and low wind speeds described more detailed in chapters \ref{solar} and \ref{low_wind}. 

\subsubsection{The Effect of Solar Irradiance} \label{solar}
Understanding how overhead lines absorb heat is crucial. Even without electricity flowing through it, the conductor's temperature rises due to solar radiation. This extra heat, added to the air temperature, is called the solar temperature. Normally, it's less than 10°C above the air temperature, but on clear nights, it can actually be lower due to heat loss. For example, if it's 35°C outside, the conductor could already be over 45°C just from the sun.  

According to \cite{Cigre601}, the heat gain per unit length is calculated using \ref{solar_general}. As seen from the equation, factors such as surface absorptivity ${\alpha_S}$, global radiation intensity ${I_T}$, and conductor diameter $D$ are affecting the heat gained from solar. The ${\alpha_S}$ can be considered as a constant, and is typically around 0.9 after a year outdoors. 
\begin{equation}
\label{solar_general}
{P_S} = {\alpha_S} * {I_T} * D
\end{equation}

In the paper, it is assumed to have always a clear sky and cloudiness is ignored as a simplification. The methodology from \cite{pvlib} is used to determine the exact solar altitude for each span by knowing time, date and location of the prediction. A constant value of 1361 W/m2 is used as the solar radiation arriving at the top of the Earth's atmosphere as suggested in \cite{solar_SHARMA1965183}, where the solar angle is affecting the results. The Fig.\ref{fig:solar_irradiance} explains solar irradiance values on a perfectly clear summer and winter solstice in Estonia. As seen, the differences are significant even with assuming the clear sky.

\Figure[t!](topskip=0pt, botskip=0pt, midskip=0pt)[width=0.99\linewidth]{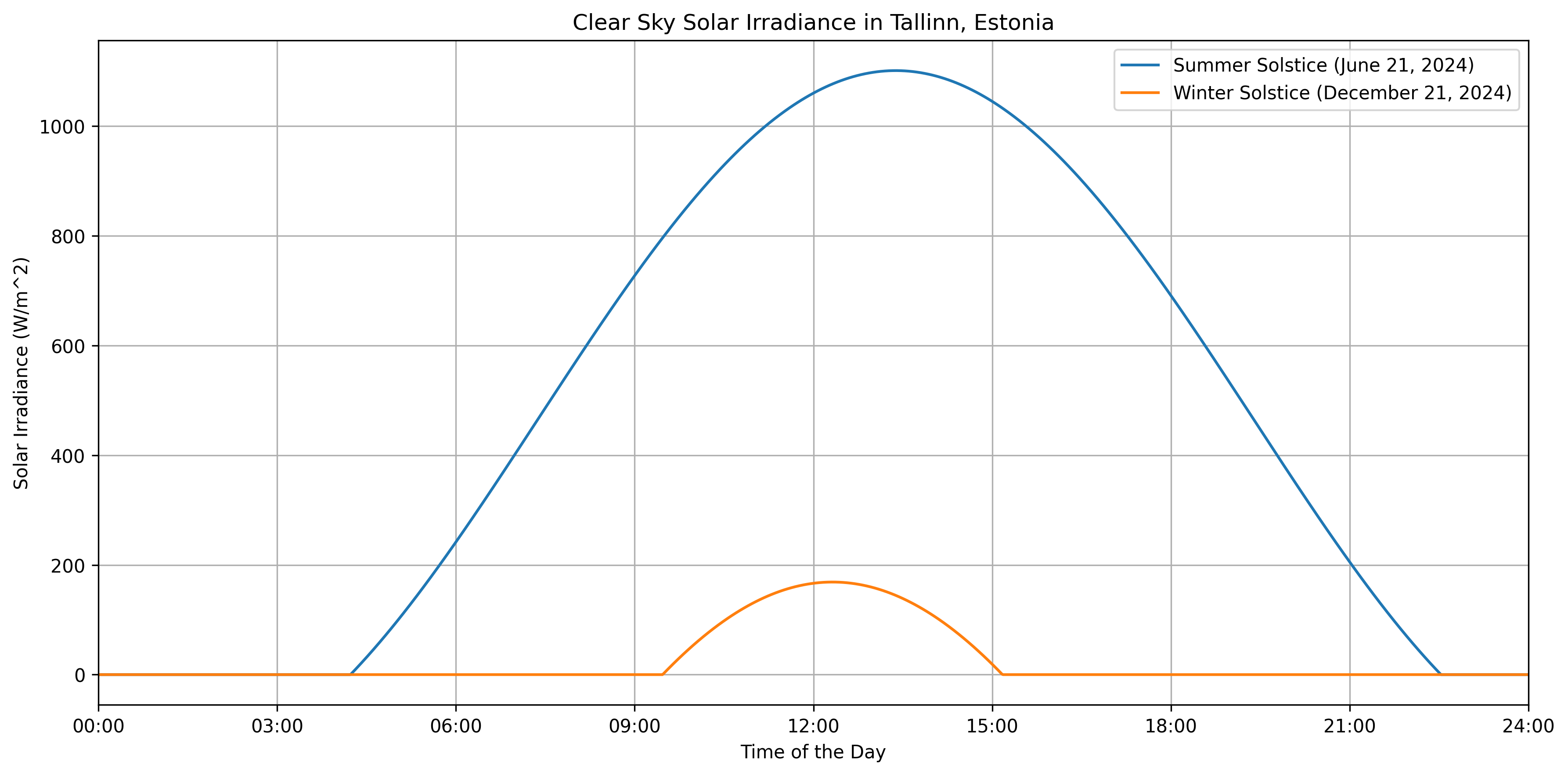}
{Solar irradiance values on a perfectly clear summer and winter solstice in Tallinn, Estonia.\label{fig:solar_irradiance}}

\subsubsection{Dealing with low wind speeds and attack angles} \label{low_wind}
Low wind speed conditions, often characterized by mean wind speeds below 2 m/s at 10 meters above ground level, hold significant implications for air pollution dispersion studies. In overhead line environments, wind direction variability is particularly notable during such periods, often appearing random. This variability challenges conventional approaches to dynamic line rating, where accurate predictions rely on stable wind conditions. In low wind speed scenarios, heat transfer physics become intricate due to buoyancy effects in the surrounding air. Various models have been proposed to account for these effects and transition smoothly between forced and natural convection. However, implementing these models in real-world situations is complex due to wind speed and direction variability over time and space.

The \cite{Cigre601} states that the wind direction in OHL environments is seldom constant, and even it can be considered essentially random at low wind speeds. Calculation reveal that parallel wind flow provides approximately 40\% less cooling than perpendicular wind flow and attack angle of 55 degrees has the cooling around 90\%. To address this variability, the concept of "effective wind speed" has been introduced for dynamic line rating as a simplification. By considering the combined effect of wind speed and direction variability, effective wind speed offers a more comprehensive assessment of cooling effects on overhead conductors. More detailed convection models for low wind speeds, based on Nusselt and Reynold numbers, are outlined in Annex C of \cite{Cigre601}, but their practical application for DLR are very complicated due to the nature of real life turbulence that can’t be easily replicated in laboratory tests.

CIGRE Technical Brochure 299 \cite{Cigre299} recommends using the higher of natural and forced convection values, as a conservative approach. This is leading to greater cooling effect on wind speed below 0.61 m/s, but still struggle at low wind speeds and angles. Additionally, CIGRE Technical Brochure 207 \cite{Cigre207} suggests, as a simplification, to use an attack angle of 45 degrees in low wind speed conditions when wind speed and wind direction standard deviations are high. These recommendations emphasize the need for accurate weather measurements and adjustments for DLR, especially in sheltered areas to ensure safety of OHLs. 

In this paper at low wind speed and attack angles, wind speed is considered to be at least 0.61m/s for the hour value and have at least 25 degrees attack angle. This is a major simplification that removes the unrealisticly low DLR values due to low wind speed and attack angle. As the inputs for the methodology are collected with 1 hour step then this simplification is providing more realistic values as the wind speed is almost never exactly 0 m/s or the attack angle is 0 for 1 hour straight. The attack angle suggestion is twice as conservative as suggested in \cite{Cigre207}.

\subsubsection{Input Parameters Time Intervals}  \label{sec:time_intervals}
The accuracy of DLR calculations is subject to uncertainties, particularly regarding the consideration of ambient parameter values over time intervals. Meteorological stations typically provide average values for 10-20 minute or even 1 hour intervals, which can introduce considerable errors in dynamic-state temperature tracking, especially as wind speed and direction are not constants in practice. DLR applications demand accurate determination of conductor temperature, especially when OHLs operate close to their thermal limits. Input parameters for these calculations must be reliable. CIGRE Technical Brochure 601 \cite{Cigre601} underscores the significance of selecting appropriate time intervals, noting that this choice significantly impacts the accuracy of conductor temperature calculations, particularly for variables like wind speed and direction that vary greatly at low speeds.

For dynamic rating applications, steady-state models are inadequate as they fail to consider the conductor's heat capacity. Rapid variations in current or weather conditions would lead to unrealistic temperature changes but also they average out peaks resulting in smoother DLR values. Weather parameters and line current are typically averaged over time intervals of 5 to 15 minutes when doing temperature tracking of OHL conductor. However, determining the average effective wind speed over such intervals can be challenging, especially if wind speed and direction change rapidly. Usually when emergency ratings are not required then it is sufficient to use longer time intervals for DLR values in steady-state. When also taking into the account the thermal capacity time periods may be even longer. In this paper weather prediction intervals are 1 hour due to the limitations of available data from weather predictions and real field measurements. As the main focus is on steady-state DLR then this approach is causing unrealistic errors, but in further research, it is recommended to use shorter time intervals with the effects of dynamic-state temperature tracking and conductor heat capacity.

\section{Case Study} \label{Sec_Casestudy} 

This section presents a case study of the proposed methodology in Estonia on an OHL that has a single weather station the middle of the OHL. The case study analyses two scenarios to get a better overview of the proposed methodology for DLR determination. For the DLR calculation solar irradiance is calculated as explained in Section \ref{solar}.

\begin{itemize}
    \item Scenario 1: all spans have the same maximum temperature of 75 degrees.
    \item Scenario 2 : the maximum temperature of each span is calculated using LiDAR measurements and sag modelling.
\end{itemize}

The case study is done in following steps:
\begin{itemize}
    \item Predict weather for all spans using model described in Section \ref{Sec:Wind}
    \item Calculate ampacity for each span and for both Scenarios as described in Section \ref{DLR}:
        \begin{enumerate}
        \item Static Rating (SR)
        \item Ambient Adjusted Rating (AAR)
        \item Dynamic Line Rating (DLR): This method has three subvariants:
                \begin{enumerate}
                \item DLR with Mean Values: Uses the predicted wind speed, direction and ambient temperature.
                \item DLR with Lower Bound: Considers the lower bound of wind speed, minimal attack angle and upper bound of ambient temperature.
                \item DLR with Upper Bound: Considers the upper bound of wind speed, angle of 90\degree and lower bound of ambient temperature.
            \end{enumerate}
        \end{enumerate}
    \item Analyse predicted and measured DLR using confidence intervals in span 44
    \item Analyse SR, AAR and DLR subvariants on a OHL level for a whole year
    \item Analyse hot-spot change in Scenario 1 and Scenario 2
\end{itemize}

\subsection{Input data} 

The selected OHL has a total of 64 spans and covers approximately 16km. The weather station is located 117m from the closest point of span 44 and has weather measurements for a period of 01.01.2023 - 31.12.2023. 

The OHL has a conductor type ACSR242 with maximum allowed temperature based on selected scenario where the subscript $temp_1$ and $temp_2$ indicate scenario. Inputs used in the case study are presented in Table. \ref{tab:case_study_data}, where three ampacity calculation methods are shown. The classical standard rating (SR), ambient adjusted rating (AAR) and the proposed DLR. CIGRE Technical Brochure 299 \cite{Cigre299} proposes to use for SR an ambient temperature close to annual or seasonal peak values, solar radiation around 1000 W/m² and an absorptivity value not less than 0.8.

\begin{table}
\centering
\caption{Description of case study inputs}
\label{tab:case_study_data}
\begin{tabularx}{\columnwidth}{|l|l|l|X|}
\hline
\textbf{Parameter}   & \textbf{SR} & \textbf{AAR} & \textbf{DLR} \\ \hline
Wind speed (m/s)           & 0.61    & 0.61     & calculated                 \\ \hline
Attack angle (\degree)     & 90      & 90       & calculated                \\ \hline
Solar irradiance (W/m2)    & 1033    & 1033     & calculated                  \\ \hline
Ambient temperature (\degree C)      &  25      & calculated & calculated            \\ \hline
Conductor type             & ACSR242 & ACSR242  & ACSR242               \\ \hline
Max $temp_1$ (\degree C)   & 75      & 75       & 75     \\ \hline
Max $temp_2 $(\degree C)   & 60      & 60       & span    \\ \hline
Solar absorption           & 0.7     & 0.7      & 0.7                \\ \hline
Solar reflectivity         & 0.7     & 0.7      & 0.7    \\ \hline
Convection coefficient     & 1       & 1        & 1    \\ \hline
\end{tabularx}
\end{table}

\subsection{Dynamic Line Rating calculation using confidence intervals} 

The use of confidence intervals (CIs) for wind speed and direction enhances the reliability of the indirect methodology by associating statistical probabilities with the most critical and variable inputs: wind speed and direction. As demonstrated in Fig.\ref{fig:DLR_span_44}, implementing CIs for wind speed, direction, and ambient temperature generally results in lower DLR values. This conservative approach ensures that the actual measured values fall within the CI bounds approximately 89\% of the time over a year, even though it falls short of the expected 95\% due to the randomness at the specific location of the weather station. Improving these results may involve more sophisticated distribution fitting or using neural networks to account for multiple distribution functions.

\Figure[t!](topskip=0pt, botskip=0pt, midskip=0pt)[width=0.99\linewidth]{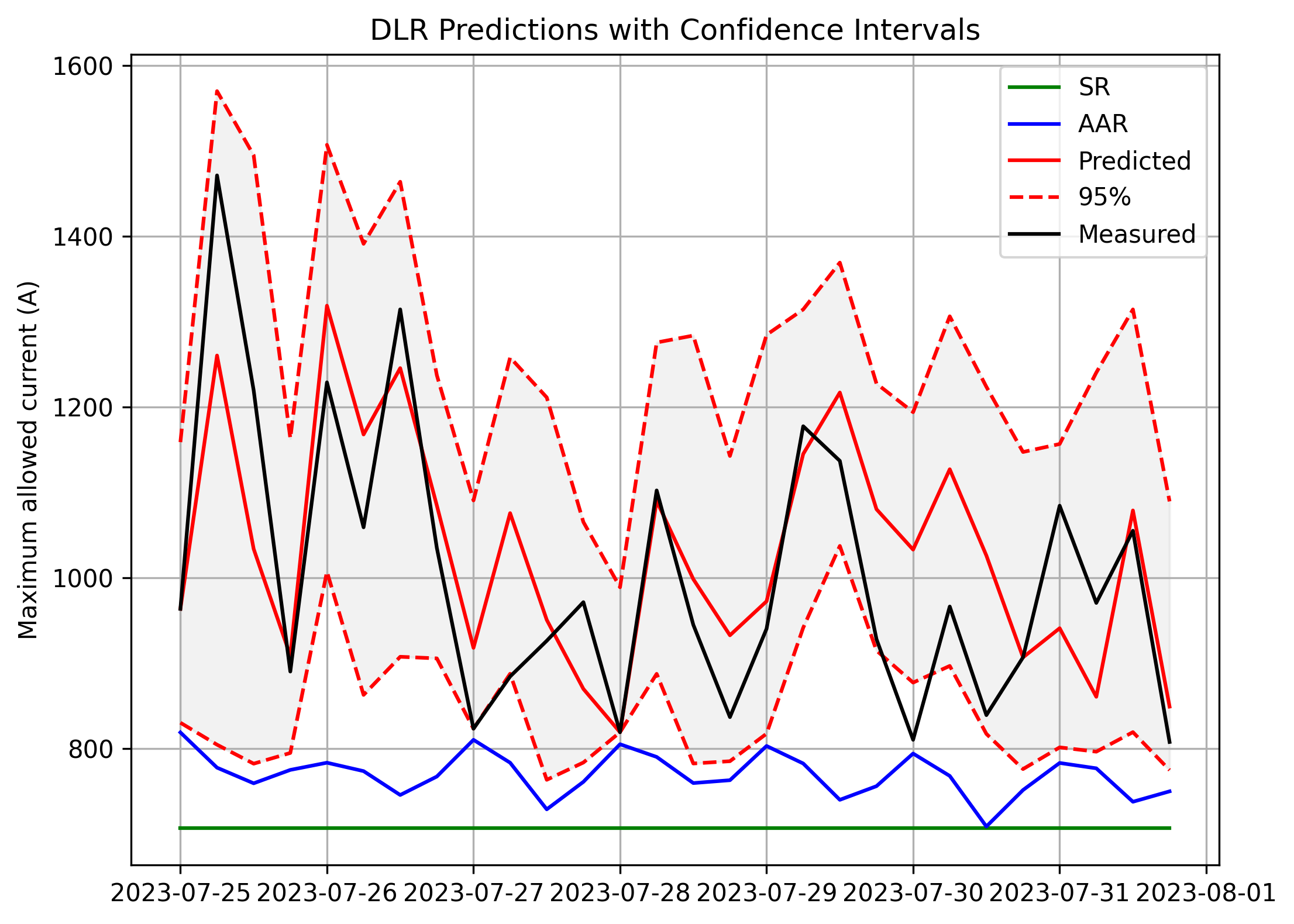}
{Dynamic Line Rating with Confidence Intervals for Span 44.\label{fig:DLR_span_44}}

Fig.\ref{DLR_1_hour} and Fig.\ref{DLR_full_year} present the results of applying the proposed methodology over a year with span-level granularity. Fig.\ref{DLR_full_year} shows how different ampacity values are calculated for each span and aggregated to determine the minimal DLR for each hour. The results indicate that the proposed methodology can predict the maximum allowed current with confidence, although the maximum DLR value is lower than when using mean values or upper bound CIs. The year-long analysis reveals that using lower bound CIs results in a year-view similar to the AAR, which uses constant values for wind inputs, and generally provides better results than the Static Rating (SR).

For OHL with varying maximum temperature spans, the methodology shows more positive results compared to OHL with uniform temperature spans. This effect is due to the better cooling conditions of the lower temperature spans, which do not cause limitations. The hotspot analysis further explains this phenomenon.

Comparing the left and right plots in Fig.\ref{DLR_full_year}, it is evident that lower bound DLR values outperform AAR for 1300 hours a year on OHL with 75-degree spans and almost all the time for OHL with different span values, as in this case study.

The results of implementing the methodology within a single time interval are shown in Fig.\ref{DLR_1_hour}, where all spans of a single OHL are displayed. Similar to Fig. \ref{DLR_full_year}, the effect of using hyper-local weather prediction is evident, producing better results when comparing DLR lower bound with AAR on OHL with varying maximum temperature spans. The left plot in Fig.\ref{DLR_full_year} shows that DLR with lower bound CI values is lower than AAR for all spans, and for spans 22 to 35, DLR with mean values is also lower than AAR, indicating poor wind conditions during that period. On the right plot, DLR with mean values is similar to DLR with lower bound on span 26, resulting in values lower than AAR.

\Figure[ht!](topskip=0pt, botskip=0pt, midskip=0pt)[width=1\linewidth]{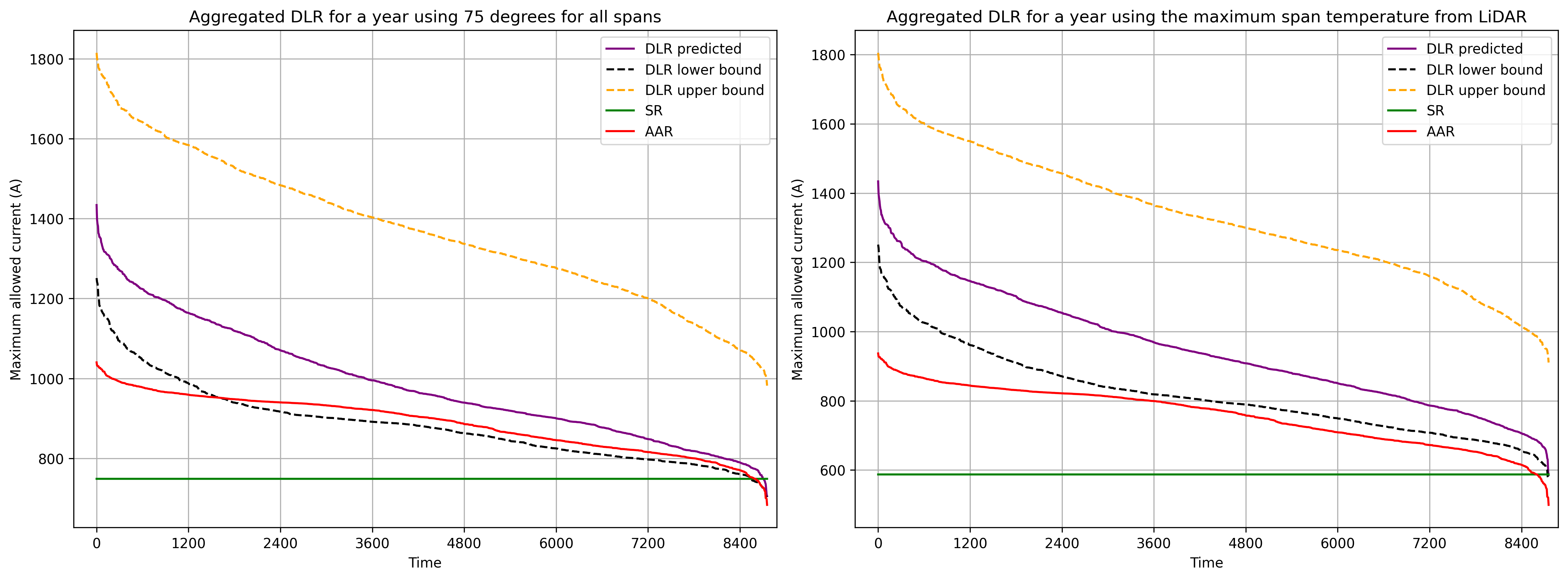}
{Dynamic Line Rating for a whole year where each hour is aggregated on minimal span value.\label{DLR_full_year}}

\Figure[ht!](topskip=0pt, botskip=0pt, midskip=0pt)[width=1\linewidth]{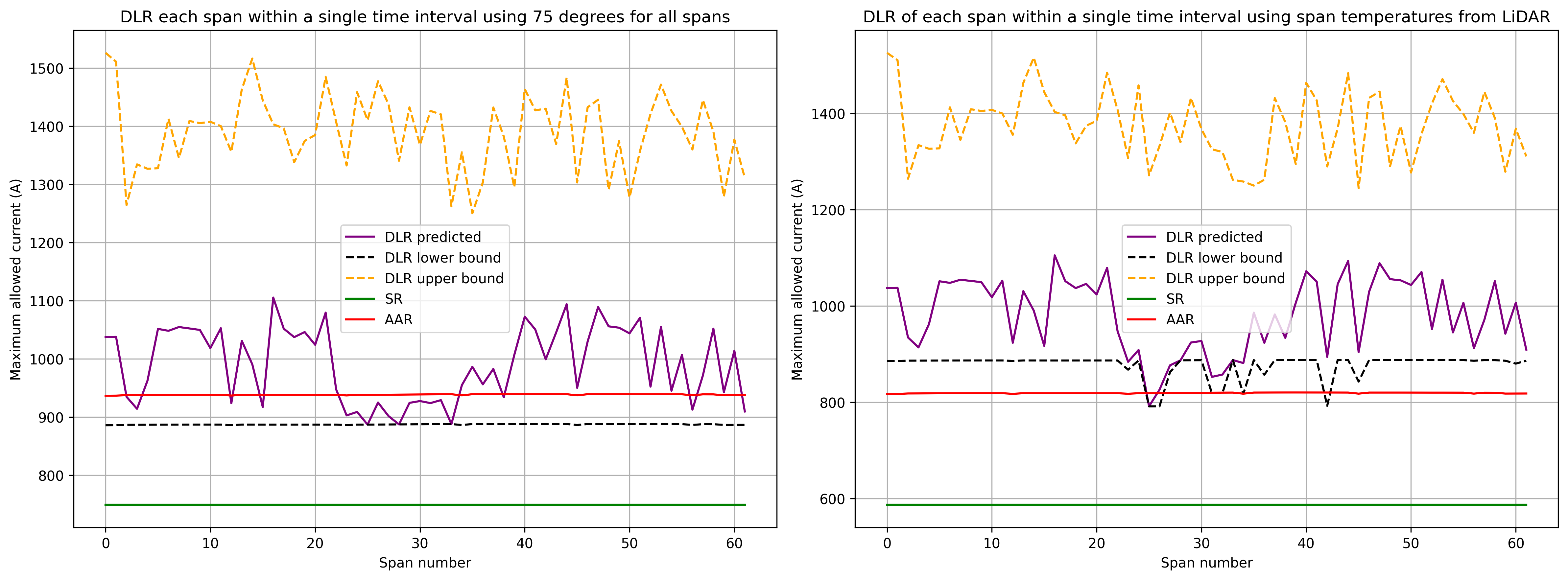}
{Dynamic Line Rating for a single hour where all spans are shown.\label{DLR_1_hour}}

\subsection{Hotspot Analysis} 

This case study analyzes the effects of implementing DLR on an OHL with both constant and variable maximum allowed temperatures for each span. The results, depicted in Fig.\ref{hotspots}, reveal significant differences in yearly capacity and hotspot distribution between these two scenarios. When the actual maximum temperature of each span is used as a limit, the overall capacity of the OHL increases compared to when all spans have a uniform maximum temperature limit. This approach considers the varying thermal limits of individual spans, allowing for more accurate and efficient utilization of the OHL's capacity.

\Figure[t!](topskip=0pt, botskip=0pt, midskip=0pt)[width=1\linewidth]{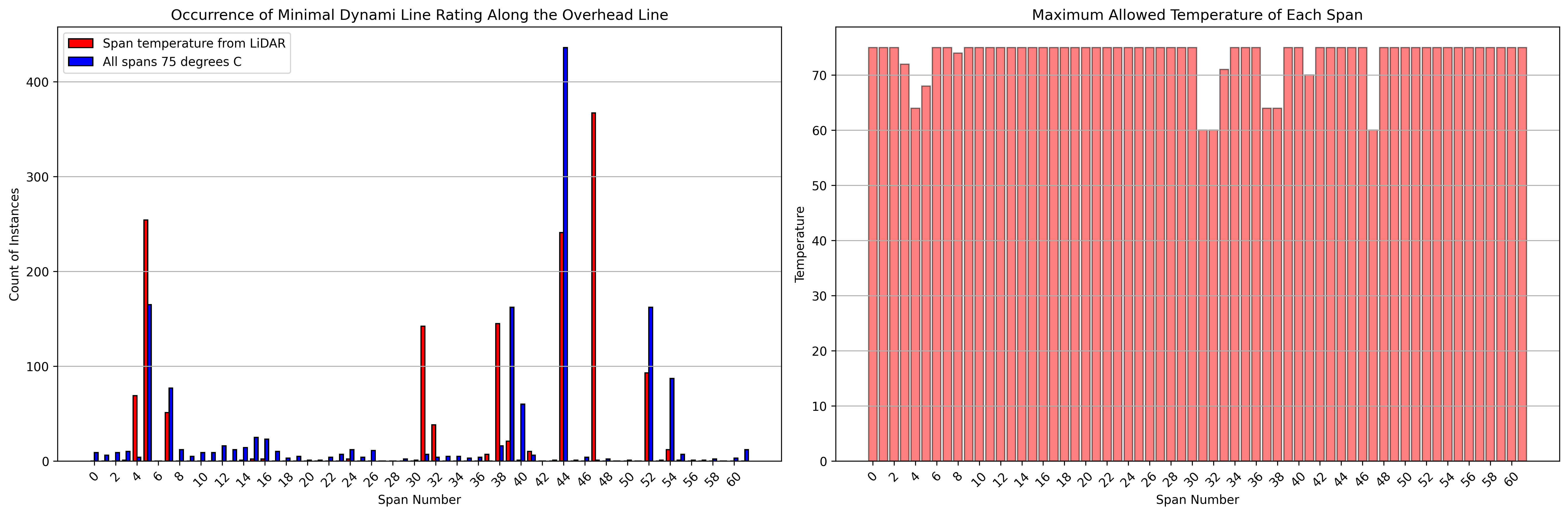}
{Critical spans change.\label{hotspots}}

The analysis shows that the limiting spans, or "hotspots," change frequently throughout the year. This variability underscores the importance of monitoring and adjusting for span-specific conditions rather than relying on a uniform temperature limit. Interestingly, even spans with maximum temperatures 15°C lower than others do not consistently limit the line's capacity. This suggests that other spans with higher maximum temperatures can also become critical under certain conditions, such as differing cooling effects from wind or ambient temperature fluctuations. By examining the year-long data, it is evident that the cooling conditions of spans with lower maximum temperatures can be advantageous. These spans do not always act as bottlenecks, as their cooler conditions allow them to handle more current without overheating. This dynamic is particularly noticeable when comparing spans within the same OHL that have different temperature limits.

The hotspot analysis highlights the benefits of a granular approach to DLR, where each span's specific conditions are taken into account. This method provides a more accurate representation of the OHL's capacity, leading to better management and optimization of the line's performance. The results indicate that using the actual maximum temperature of each span can significantly enhance the OHL's capacity compared to a uniform maximum temperature approach. Additionally, the changing nature of hotspots throughout the year suggests that a static rating system is less effective in managing OHL performance. A dynamic approach, which continuously adapts to real-time data and conditions, offers superior results by identifying and addressing the most critical spans at any given time.

This study also demonstrates the importance of comprehensive monitoring and data collection across all spans of an OHL. By understanding the unique characteristics and behavior of each span, operators can implement more effective DLR strategies that enhance the reliability and efficiency of the power transmission system. Overall, the hotspot analysis confirms that using span-specific maximum temperatures and real-time data significantly improves the OHL's capacity and performance. This approach ensures that the line operates within safe thermal limits while maximizing its ability to transmit electricity efficiently, ultimately leading to a more robust and reliable power grid.

\subsection{Hyper-local Dynami Line Rating in Practise} 

An example from the selected OHL in Fig.\ref{map_with_wind} illustrates how specific DLR values are calculated for each span based on input parameters. Wind speed and direction are shown as arrows, with the arrow length representing wind speed and the arrow direction indicating the predicted wind direction. The green arrow represents the measured wind speed and direction near the weather station next to span 44. This example clearly demonstrates the variability in wind conditions along the OHL, causing different spans to limit DLR at different times.

\Figure[t!](topskip=0pt, botskip=0pt, midskip=0pt)[width=1\linewidth]{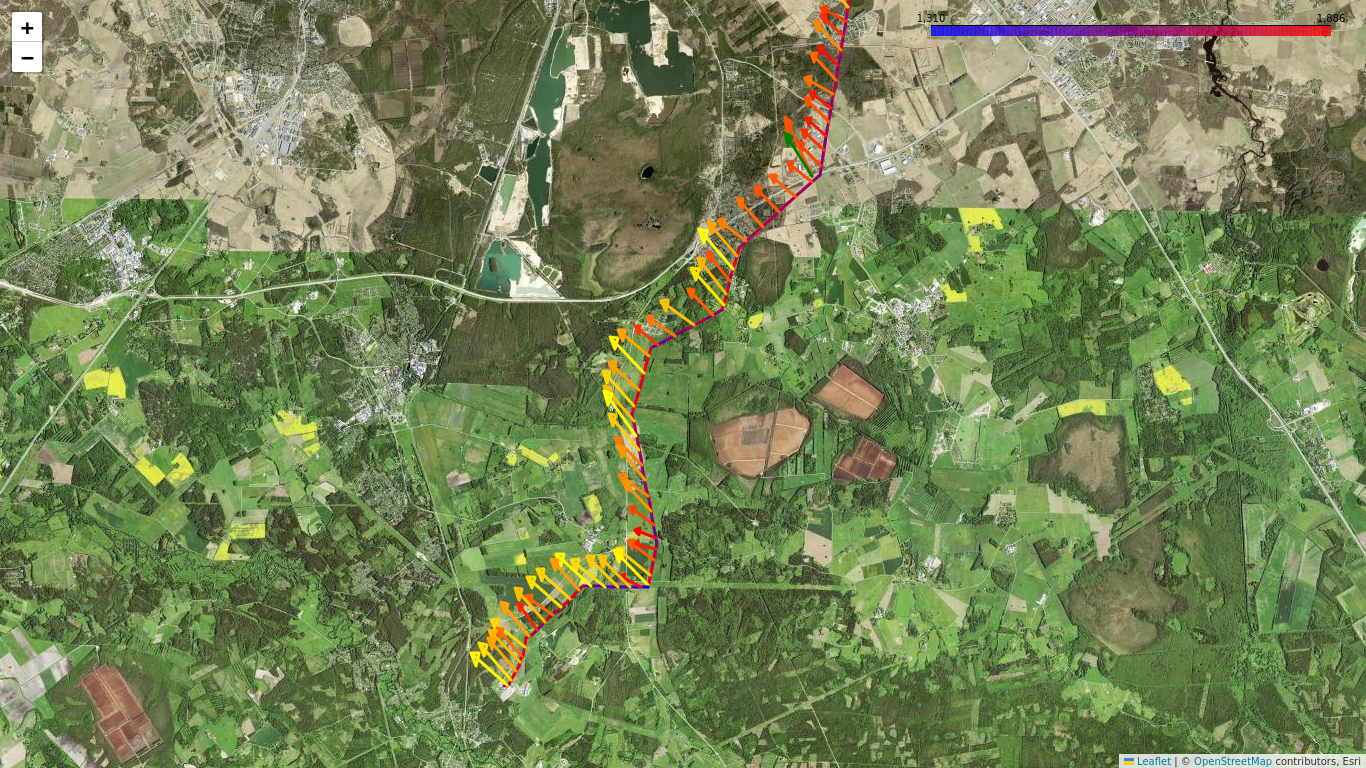}
{Calculated Dynamic Line Rating With Predicted Wind for Each Span.\label{map_with_wind}}

\section{Results} 
This paper presents a comprehensive study on the implementation of Dynamic Line Rating (DLR) using confidence intervals (CIs) for wind speed, wind direction, and ambient temperature. The methodology aims to increase the reliability of indirect DLR calculations by incorporating the hyper-local weather prediction model together with statistical probabilities associated with these key inputs. The results demonstrate the effectiveness of this approach in predicting the maximum ampacity of OHLs more accurately than traditional methods.

The study analyzed the impact of using CIs on DLR values over a year-long period and within specific time intervals. It was found that incorporating CIs reduces the effective DLR value, providing a more conservative and reliable estimate with certainty that is essential for system operators. The analysis showed that 89\% of the DLR values based on measured data fell within the predicted CI bounds for a specific weather station, albeit slightly lower than the expected 95\% due to the randomness and novelty of the location for the weather prediction model. To enhance the accuracy, more sophisticated distribution fitting methods and neural networks with more training data in various environments are suggested.

The research also examined the effects of applying DLR at a span-level granularity for an entire year, comparing different spans within a single hour, and analyzing the impact of hyper-local weather conditions. It was observed that using the lower bound of CIs provided better results than using average values or upper bounds, aligning closely with the performance of Ambient Adjusted Rating (AAR) in many cases. The methodology proved particularly effective for OHLs with spans that have varying maximum allowable temperatures, showing improved capacity utilization and dynamic adaptability to changing weather conditions.

\section{Conclusions and further works} 

In conclusion, this paper demonstrates that incorporating confidence intervals into DLR calculations provides a more reliable, efficient, and adaptable method for managing the capacity of overhead lines. The findings underscore the importance of detailed, span-level analysis and real-time data integration in optimizing the performance of power transmission systems. This approach represents a significant advancement in the field of power grid management, offering a robust solution for enhancing the reliability and efficiency of electricity transmission.

Future work should focus on improving the accuracy of DLR calculations by incorporating more advanced distribution fitting methods and neural network models. These enhancements will help achieve the expected 95\% or 99.7\% confidence bounds and further increase the reliability of DLR predictions. Having more reliable prediction results also gives better CI bounds resulting in larger DLR values as lower bound is rising. Integrating shorter than 1 hour time intervals or even real-time weather data and enhancing hyper-local weather prediction models will also improve the dynamic adaptability of DLR calculations, especially when dealing with values that are close to the thermal limit of OHLs.

Overall, the proposed methodology represents a significant advancement in DLR prediction, offering a robust solution for optimizing the performance and reliability of electricity transmission systems without adding sensors to high voltage lines.

\bibliographystyle{IEEEtran}
\bibliography{bibliography.bib}

\EOD

\end{document}